%% file: main.tex
\relax
\documentclass[letterpaper]{article} %DO NOT CHANGE THIS
\usepackage{conll-2019}  %Required
\usepackage{times}  %Required

\usepackage{latexsym}
\usepackage{amsmath}
\usepackage{amssymb}
\usepackage{mathtools}
\usepackage{enumitem}
\usepackage{multirow}
\usepackage{url}
\usepackage{bm}
\usepackage{tabularx}
\usepackage[normalem]{ulem}

\usepackage{algpseudocode}

\algnewcommand\algorithmicDefine{\textbf{Define:}}
\algnewcommand\Define{\item[\algorithmicDefine]}


\newcommand{\y}{\mathbf{y}}
\newcommand{\x}{\mathbf{x}}
\newcommand{\G}{\mathbf{G}}
\newcommand{\loss}{\mathcal{L}}

\usepackage{booktabs}

\usepackage{amsmath}
\usepackage{graphicx}
\aclfinalcopy % Uncomment this line for the final submission
\def\w{\textbf{w}}
\def\x{\textbf{x}}

\definecolor{LinkColor}{RGB}{0,0,130}

\title{Named Entity Recognition with Partially Annotated Training Data}

\author{Stephen Mayhew$^\natural$, Snigdha Chaturvedi$^\sharp$, Chen-Tse Tsai$^\flat$, Dan Roth$^\natural$\\
	    $^\natural$University of Pennsylvania, Philadelphia, PA, 19104\\
	    $^\sharp$University of North Carolina, Chapel Hill, NC, 27599\\
	    $^\flat$Bloomberg LP\\
	    {\tt \{mayhew, danroth\}@seas.upenn.edu},\\ {\tt snigdha@cs.unc.edu, ctsai54@bloomberg.net}}

\begin{document}
\maketitle

\begin{abstract}
Supervised machine learning assumes the availability of fully-labeled data, but in many cases, such as low-resource languages, the only data available is partially annotated. We study the problem of Named Entity Recognition (NER) with partially annotated training data in which a fraction of the named entities are labeled, and all other tokens, entities or otherwise, are labeled as non-entity by default. In order to train on this noisy dataset, we need to distinguish between the true and false negatives. To this end, we introduce a constraint-driven iterative algorithm that learns to detect false negatives in the noisy set and downweigh them, resulting in a weighted training set. With this set, we train a weighted NER model. We evaluate our algorithm with weighted variants of neural and non-neural NER models on data in 8 languages from several language and script families, showing strong ability to learn from partial data. Finally, to show real-world efficacy, we evaluate on a Bengali NER corpus annotated by non-speakers, outperforming the prior state-of-the-art by over 5 points F1.
\end{abstract}

\section{Introduction}

Most modern approaches to NLP tasks rely on supervised learning algorithms to learn and generalize from labeled training data. While this has proven successful in high-resource scenarios, this is not realistic in many cases, such as low-resource languages, as the required amount of training data just doesn't exist. However, partial annotations are often easy to gather.

\begin{figure}[t]
\centering
\includegraphics[scale=0.16]{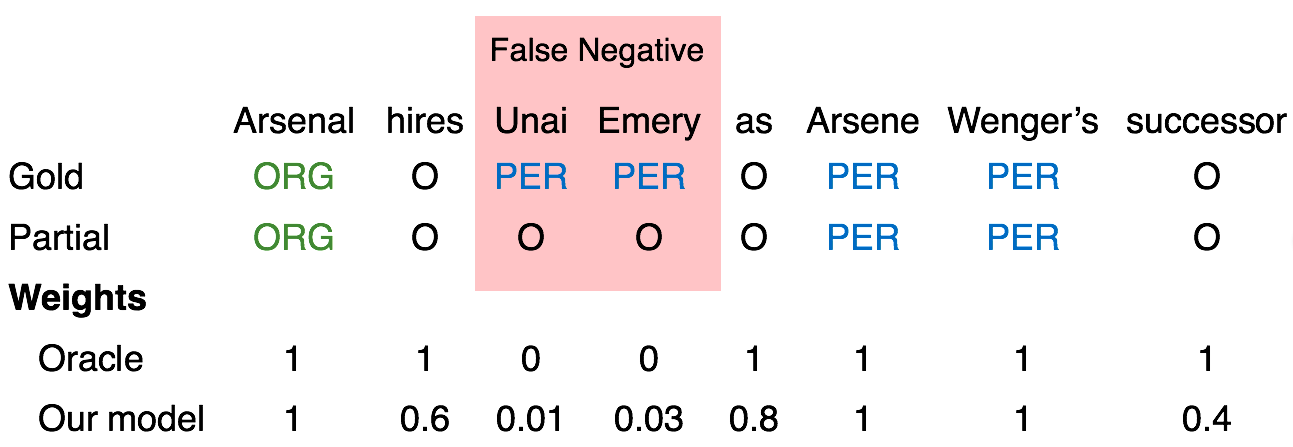}
\caption{This example has three entities: \textit{Arsenal}, \textit{Unai Emery}, and \textit{Arsene Wenger}. In the \textit{Partial} row, the situation addressed in this paper, only the first and last are tagged, and all other tokens are assumed to be non-entities, making \textit{Unai Emery} a false negative as compared to \textit{Gold}. Our model is an iteratively learned binary classifier used to assign weights to each token indicating its chances of being correctly labeled. The \textit{Oracle} row shows optimal weights.}
\label{example}
\end{figure}

We study the problem of using partial annotations to train a Named Entity Recognition (NER) system. In this setting, all (or most) identified entities are correct, but not all entities have been identified, and crucially, there are no reliable examples of the negative class. The sentence shown in Figure~\ref{example} shows examples of both a gold and a partially annotated sentence. Such partially annotated data is relatively easy to obtain: for example, a human annotator who does not speak the target language may recognize common entities, but not uncommon ones. With no reliable examples of the negative class, the problem becomes one of estimating which unlabeled instances are true negatives and which are false negatives.  

To address the above-mentioned challenge, we present Constrained Binary Learning (CBL) -- a novel self-training based algorithm that focuses on iteratively identifying true negatives for the NER task while improving its learning. Towards this end, CBL uses constraints that incorporate background knowledge required for the entity recognition task.

We evaluate the proposed methods in 8 languages, showing a significant ability to learn from partial data. We additionally experiment with initializing CBL with domain-specific instance-weighting schemes, showing mixed results. In the process, we use weighted variants of popular NER models, showing strong performance in both non-neural and neural settings. Finally, we show experiments in a real-world setting, by employing non-speakers to manually annotate romanized Bengali text. We show that a small amount of non-speaker annotation combined with our method can outperform previous methods. 

\section{Related Work}

The supervision paradigm in this paper, partial supervision, falls broadly under the category of semi-supervision \cite{chapelle2009semi}, and is closely related to weak supervision \cite{hernandez2016weak}\footnote{See also: \url{https://hazyresearch.github.io/snorkel/blog/ws_blog_post.html}} and incidental supervision \cite{Roth17}, in the sense that data is constructed through some noisy process. However, all of the most related work shares a key difference from ours: reliance on a small amount of fully  annotated data in addition to the noisy data.

\newcite{FernandesBr11} introduces a transductive version of structured perceptron for partially annotated sequences. However, their definition of partial annotation is labels removed at random, so examples from all classes are still available if not contiguous.
 
Fidelity Weighted Learning \cite{DMGKS17} uses a teacher/student model, in which the teacher has access to (a small amount) of high quality data, and uses this to guide the student, which has access to (a large amount) of weak data. 

\newcite{HedderichKl18}, following \newcite{GoldbergerBe17}, add a noise adaptation layer on top of an LSTM, which learns how to correct noisy labels, given a small amount of training data. We compare against this model in our experiments.

In the world of weak supervision, Snorkel \cite{RBEFWR17,FWRR17}, is a system that combines automatic labeling functions with data integration and noise reduction methods to rapidly build large datasets. They rely on high recall and consequent redundancy of the labeling functions. We argue that in certain realistic cases, high-recall candidate identification is unavailable.

We draw inspiration from the Positive-Unlabeled (PU) learning framework \cite{LLYL02,LDLLY03,LeeLi03,ElkanNo08}. Originally introduced for document classification, PU learning addresses problems where examples of a single class (for example, sports) are easy to obtain, but a full labeling of all other classes is prohibitively expensive. 

Named entity classification as an instance of PU learning was introduced in \newcite{Grave14}, which uses constrained optimization with constraints similar to ours.  However, they only address the problem of named entity classification, in which mentions are given, and the goal is to assign a \emph{type} to a named-entity (like `location', `person', etc.) as opposed to our goal of identifying and typing named entities.

Although the task is slightly different, there has been work on building `silver standard' data from Wikipedia \cite{NothmanCuMa08,NRRMC13,PZMNKJ17}, using hyperlink annotations as the seed set and propagating throughout the document.

Partial annotation in various forms has also been studied in the contexts of POS-tagging \cite{MNNS15}, word sense disambiguation \cite{HovyHo12}, temporal relation extraction \cite{NYFD18}, dependency parsing \cite{FMNM12}, and named entity recognition \cite{JXLDL19}.

In particular, \citet{JXLDL19} study a similar problem with a few key differences: since they remove entity surfaces randomly, the dataset is too easy; and they do not use constraints on their output. We compare against their results in our experiments.

Our proposed method is most closely aligned with the Constraint Driven Learning (CoDL) framework \cite{ChangRaRo07}, in which an iterative algorithm reminiscent of self-training is guided by constraints that are applied at each iteration. 

\section{Constrained Binary Learning}
\label{sec:binary}

Our method assigns instance weights to all negative elements (tokens tagged as O), so that false negatives have low weights, and all other instances have high weights. We calculate weights according to the confidence predictions of a classifier trained iteratively over the partially annotated data. We refer to our method as Constrained Binary Learning (CBL).\footnote{Publication details (including code) can be found at \mbox{\url{cogcomp.org/page/publication_view/888}}}

We will first describe the motivation for this approach before moving on to the mechanics. We start with partially annotated data (which we call set $T$) in which some, but not all, positives are annotated (set $P$), and no negative is labeled. By default, we assume that any instance not labeled as positive is labeled as negative as opposed to unlabeled. This data (set $N$) is noisy in the sense that many true positives are labeled as negative (these are \textit{false negatives}). Clearly, training on $T$ as-is will result in a noisy classifier.

Two possible approaches are: \textbf{1)} find the false negatives and label them correctly, or \textbf{2)} find the false negatives and remove them. The former method affords more training data, but runs the risk of adding noise, which could be worse than the original partial annotations. The latter is more forgiving because of an asymmetry in the penalties: it is important to remove all false negatives in $N$, but inadvertently removing true negatives from $N$ is typically not a problem, especially in NER, where negative examples dominate. Further, a binary model (only two labels) is sufficient in this case, as we need only detect entities, not type them.

We choose the latter method, but instead of removing false negatives, we adopt an instance-weighting approach, in which each instance is assigned a weight $v_i \geq 0$ according to confidence in the labeling of that instance. A weight of $0$ means that the loss this instance incurs during training will not update the model.

With this in mind, CBL takes two phases: first, it learns a binary classifier $\lambda$ using a constrained iterative process modeled after the CODL framework \cite{ChangRaRo07}, and depicted in Figure \ref{algorithm}. The core of the algorithm is the train-predict-infer loop. The training process (line 4) is weighted, using weights $V$. At the start, these can be all 1 (Raw), or can be initialized with prior knowledge. The learned model is then used to predict on all of $T$ (line 5). In the inference step (line 6), we take the predictions from the prior round and the constraints $C$ and produce a new labeling on $T$, and a new set of weights $V$. The details of this inference step are presented later in this section. Although our ultimate strategy is simply to assign weights (not change labels), in this inner loop, we update the labels on $N$ according to classifier predictions.

\begin{figure}
\begin{algorithmic}[1]
%\Define
%\Statex 1, 0: Positive and negative labels
\Require
\Statex $P$ : positive tokens
\Statex $N$ : noisy negative tokens
\Statex $C$ : constraints
\Statex
\State $T = N \cup P$
\State $V \gets $ Initialize $T$ with weights (Optional)
\While{stopping condition not met}
	\State $\lambda \gets \text{train}(T, V)$	
	\State $\hat{T} \gets \text{predict}(\lambda, T)$
  	\State $T,V \gets \text{inference}(\hat{T}, C)$ 
\EndWhile
%\State $W \gets \text{getWeights}(T)$
\State return $\lambda$
\end{algorithmic}
\caption{Constrained Binary Learning (CBL) algorithm (phase 1). The core of the algorithm is in the while loop, which iterates over training on $T$, predicting on $T$ and correcting those predictions.}
\label{algorithm}
\end{figure}

In the second phase of CBL, we use the $\lambda$ trained in the previous phase to assign weights to instances as follows: \begin{equation}
 \label{eqn:weightassign}
   v_i =
   \begin{cases}
     1.0 & \text{if $x_i \in P$} \\
     P_{\lambda}(y_i=\text{O} \mid x_i) & \text{if $x_i \in N$} \\
   \end{cases}
 \end{equation}

Where $P_{\lambda}(y_i=\text{O} \mid x_i)$ is understood as the classifier's confidence that instance $x_i$ takes the negative label. In practice it is sufficient to use any confidence score from the classifier, not necessarily a probability. If the classifier has accurately learned to detect entities, then for all the false negatives in $N$, $P_{\lambda}(y_i=\text{O}|x_i)$ is small, which is the goal. 

Ultimately, we send the original multiclass partially annotated dataset along with final weights $V$ to a standard weighted NER classifier to learn a model. No weights are needed at test time.

\subsection{NER with CBL}
So far, we have given a high-level view of the algorithm. In this section, we will give more low-level details, especially as they relate to the specific problem of NER. One contribution of this work is the inference step (line 6), which we address using a constrained Integer Linear Program (ILP) and describe in this section. However, the constraints are based on a value we call the \textit{entity ratio}. First, we describe the entity ratio, then we describe the constraints and stopping condition of the algorithm.

\subsubsection{Entity ratio and Balancing}

We have observed that NER datasets tend to hold a relatively stable ratio of entity tokens to total tokens. We refer to this ratio as $b$, and define it with respect to some labeled dataset as: 
\begin{equation}
b = \frac{|P|}{|P| + |N| }
\label{eqn:entitymention}
\end{equation}

where $N$ is the set of negative examples. Previous work has shown that in fully-annotated datasets the entity ratio tends to be about $0.09 \pm 0.05$, depending on the dataset and genre \cite{AugensteinDeBo2017}. Intuitively, knowledge of the gold entity ratio can help us estimate when we have found all the false negatives.

In our main experiments, we assume that the entity ratio with respect to the gold labeling is known for each training dataset. A similar assumption was made in \newcite{ElkanNo08} when determining the $c$ value, and in \newcite{Grave14} in the constraint determining the percentage of {\sc other} examples. However, we also show in Section \ref{sec:bratio} that knowledge of this ratio is not strictly necessary, and a flat value across all datasets produces similar performance.

With a weighted training set, it is also useful to define the weighted entity ratio.

\begin{equation}
b = \frac{|P|}{|P| + \sum_{i \in N} v_i }
\label{eqn:weightedmention}
\end{equation}

When training an NER model on weighted data, one can change the weighted entity ratio to achieve different effects. To make balanced predictions on test, the entity ratio in the training data should roughly match that of the test data \cite{Chawla05}. To bias a model towards predicting positives or predicting negatives, the weighted entity ratio can be set higher or lower respectively. This effect is pronounced when using linear methods for NER, but not as clear in neural methods.

To change the entity ratio, we scale the weights in $N$ by a scaling constant $\gamma$. Targeting a particular $b^*$, we may write:
\begin{equation}
b^* = \frac{|P|}{|P| + \gamma \sum_{i \in N} v_i }
\end{equation}

We can solve for $\gamma$:
\begin{equation}
\gamma = \frac{(1-b^*)|P|}{b^* \sum_{i \in N} v_i}
\label{eqn:gamma}
\end{equation}

To obtain weights, $v^*_i$, that attain the desired entity ratio, $b^*$, we scale all weights in $N$ by $\gamma$.
\begin{equation}
  v^*_i = \gamma v_i
\end{equation}

In the train-predict-infer loop, we balance the weights to a value near the gold ratio before training.

\subsubsection{Constraints and Stopping Condition}

We encode our constraints with an Integer Linear Program (ILP), shown in Figure \ref{binaryilp}. Intuitively, the job of the inference step is to take predictions ($\hat{T}$) and use knowledge of the task to `fix' them. 

In the objective function (Eqn. \ref{objective}), token $i$ is represented by two indicator variables $y_{0i}$ and $y_{1i}$, representing negative and positive labels, respectively. Associated prediction scores $C_0$ and $C_1$ are from the classifier $\lambda$ in the last round of predictions. The first constraint (Eqn. \ref{cantbothbeone}) encodes the fact that an instance cannot be both an entity and a non-entity.

The second constraint (Eqn. \ref{posconstraint}) enforces the ratio of positive to total tokens in the corpus to match a required entity ratio. $|T|$ is the total number of tokens in the corpus. $b$ is the required entity ratio, which increases at each iteration. $\delta$ allows some flexibility, but is small.

Constraint \ref{pconstraint} encodes that instances in $P$ should be labeled positive since they were manually labeled and are by definition trustworthy. We set $\xi \geq 0.99$.

This framework is flexible in that more complex language- or task-specific constraints could be added. For example, in English and many other languages with Latin script, it may help to add a capitalization constraint. In languages with rich morphology, certain suffixes may indicate or contraindicate a named entity. For simplicity, and because of the number of languages in our experiments, we use only a few constraints. 

After the ILP has selected predictions, we assign weights to each instance in preparation for training the next round. The decision process for an instance is:
 \begin{equation}
   v_i =
   \begin{cases}
     1.0 & \text{If ILP labeled $x_i$ positive} \\
     P_{\lambda}(y_i=\text{O} \mid x_i) & \text{Otherwise} \\
   \end{cases}
 \end{equation}

This is similar to Equation (\ref{eqn:weightassign}), except that the set of tokens that the ILP labeled as positive is larger than $P$. With new labels and weights, we start the next iteration.

The stopping condition for the algorithm is related to the entity ratio. One important constraint (Eqn. \ref{posconstraint}) governs how many positives are labeled at each round. This number starts at $|P|$ and is increased by a small value\footnote{The size of this value is related to how much we trust the ranking induced by prediction confidences. If we believed the ranking was perfect, we could take as many positives as we wanted and be finished in one round.} at each iteration, thereby improving recall. Positive instances are chosen in two ways. First, all instances in $P$ are constrained to be labeled positive (Eqn. \ref{pconstraint}). Second, the objective function ensures that high-confidence positives will be chosen. The stopping condition is met when the number of required positive instances (computed using gold unweighted entity ratio) equals the number of predicted positive instances.

\begin{figure}
\begin{align}
& \underset{\textbf{y}}{\text{max}} & & \sum_i^{|T|} C_{0i} y_{0i} + C_{1i} y_{1i} \label{objective}  \\
& \text{s.t.} & & \forall i,~ y_{0i} + y_{1i} = 1 \label{cantbothbeone} \\
& & & b-\delta \leq \sum_i y_{1i} / |T| \leq b+\delta \label{posconstraint} \\
%& & & \forall i,  \text{isPuncNum}(x_i), \sum_i y_{0i} \geq \epsilon |\text{PuncNum}|,  \\
& & & \forall i,~ x_i \in P,~ \sum_i y_{1i} \geq \xi |P|,  \label{pconstraint}
% &&& \forall s \in S, \label{consista}\\
% & && ~|s|a_s + \sum_{i \in s} y_{1i} \geq \alpha|s| \\
% & && ~|s|b_s + \sum_{i \in s} y_{0i} \geq \alpha|s| \\
% & && ~a_s+b_s=1 \label{consistb}
\end{align}
\caption{ILP for the inference step}
\label{binaryilp}
\end{figure}

\section{Experiments}
\label{sec:experiments}

We measure the performance of our method on 8 different languages using artificially perturbed labels to simulate the partial annotation setting.

\subsection{Data}
We experiment on 8 languages. Four languages -- English, German, Spanish, Dutch -- come from the CoNLL 2002/2003 shared tasks \cite{TjongKimSang02,TjongKimSangMe03}. These are taken from newswire text, and have labelset of Person, Organization, Location, Miscellaneous.

The remaining four languages come from the LORELEI project \cite{StrasselTr16}. These languages are: Amharic (amh: LDC2016E87), Arabic (ara: LDC2016E89), Hindi (hin: LDC2017E62), and Somali (som: LDC2016E91). These come from a variety of sources including discussion forums, newswire, and social media. The labelset is Person, Organization, Location, Geo-political entity. We define train/development/test splits, taking care to keep a similar distribution of genres in each split. Data statistics for all languages are shown in Table \ref{tab:datastats}.

\subsection{Artificial Perturbation}
We create partial annotations by perturbing gold annotated data in two ways: lowering recall (to simulate missing entities), and lowering precision (to simulate noisy annotations).

To lower recall, we replace gold named entity tags with $O$ tags (for non-name). We do this by grouping named entity surface forms, and replacing tags on all occurrences of a randomly selected surface form until the desired amount remains. For example, if the token `Bangor' is chosen to be untagged, then every occurrence of `Bangor' will be untagged. We chose this slightly complicated method because the simplest idea (remove mentions randomly) leaves an artificially large diversity of surface forms, which makes the problem of discovering noisy entities easier.

To lower precision, we tag a random span (of a random start position, and a random length between $1$ and $3$) with a random named entity tag. We continue this process until we reach the desired precision. When both precision and recall are to be perturbed, the recall adjustment is made first, and then the number of random spans to be added is calculated by the entities that are left.

\subsection{NER Models}

In principle, CBL can use any NER method that can be trained with instance weights. We experiment with both non-neural and neural models.

\subsubsection{Non-neural Model}

For our non-neural system, we use a version of Cogcomp NER \cite{RatinovRo09,KSZRCSRRLDTRMFWYSGUANLR18} modified to use Weighted Averaged Perceptron. This operates on a weighted training set $D_w = \{ (x_i, y_i, v_i) \}_{i=1}^N $, where $N$ is the number of training examples, and $v_i \geq 0$ is the weight on the $i$th training example. In this non-neural system, a training example is a word with context encoded in the features. We change only the update rule, where the learning rate $\alpha$ is multiplied by the weight:\begin{equation}
\w = \w + \alpha v_i y_i (\w^T x_i)
\end{equation}

We use a standard set of features, as documented in \citet{RatinovRo09}. In order to keep the language-specific resources to a minimum, we did not use any gazetteers for any language.\footnote{Separate experiments show that omitting gazetteers impacts performance only slightly.} One of the most important features is Brown clusters, trained for 100, 500, and 1000 clusters for the CoNLL languages, and 2000 clusters for the remaining languages. We trained these clusters on Wikipedia text for the four CoNLL languages, and on the same monolingual text used to train the word vectors (described in Section \ref{sec:neural}).

\begin{table}[t!]
\small
  \begin{center}
    \begin{tabular}{lrrrrrr}
    \toprule
    & \multicolumn{3}{c}{Train} & \multicolumn{3}{c}{Test} \\
    \cmidrule(r){2-4} \cmidrule(l){5-7}
    Lang. & $b$ (\%) & Tag & Tok & $b$ (\%) & Tag & Tok \\
    \midrule
    English & 16.6 & 23K & 203K & 17.3 & 5K & 46K \\
    Spanish & 12.3 & 18K & 264K & 11.9 & 3K & 51K \\
    German & 8.0 & 11K & 206K & 9.9 & 3K & 51K \\
    Dutch & 9.5 & 13K & 202K & 8.3 & 4K & 68K \\
    Amharic & 11.2 & 3K & 52K & 11.3 & 1K & 18K \\
    Arabic & 12.6 & 4K & 60K & 10.2 & 931 & 16K \\
    Hindi & 7.38 & 4K & 74K & 7.53 & 1K & 25K \\
    Somali & 11.2 & 4K & 57K & 11.9 & 1K & 16K \\
    \bottomrule
    \end{tabular}
  \end{center}
  \caption{Data statistics for all languages, showing number of tags and tokens in Train and Test. The tag counts represent individual spans, not tokens. That is, ``$[\text{Barack Obama}]_{\text{PER}}$'' counts as one tag, not two. The $b$ column shows the entity ratio as a percentage.}
\label{tab:datastats}
\end{table}

\subsubsection{Neural Model}
\label{sec:neural}
A common neural model for NER is the BiLSTM-CRF model \cite{ma2016end}. However, because the Conditional Random Field (CRF) layer calculates loss at the sentence level, we need a different method to incorporate token weights. We use a variant of the CRF that allows partial annotations by marginalizing over all possible sequences \cite{TKOMM08}.

When using a standard BiLSTM-CRF model, the loss of a dataset ($D$) composed of sentences ($s$) is calculated as:

\begin{equation}
\loss = - \sum_{s\in D} \log P_\theta (\y^{(s)} | \x^{(s)})
\end{equation}

Where $P_\theta(\y^{(s)} | \x^{(s)})$ is calculated by the CRF over outputs from the BiLSTM. In the marginal CRF framework, it is assumed that $\y^{(s)}$ is necessarily partial, denoted as $\y^{(s)}_p$. To incorporate partial annotations, the loss is calculated by marginalizing over all possible sequences consistent with the partial annotations, denoted as $C(\y_p^s)$. 

\begin{equation}
\label{eq:tsuboi}
\loss = - \sum_{s\in D} \log \sum_{\y \in C(\y^{(s)}_p)} P_\theta(\y | \x^{(s)})    
\end{equation}

However, this formulation assumes that all possible sequences are equally likely. To address this, \citet{JXLDL19} introduced a way to weigh sequences. 

\begin{equation}
\loss = - \sum_{s\in D} \log \sum_{\y \in C(\y^{(s)}_p)} q(\y|\x^{(s)}) P_\theta(\y | \x^{(s)})
\end{equation}

It's easy to see that this formulation is a generalization of the standard CRF if $q(.)=1$ for the gold sequence $\y$, and 0 for all others.

The product inside the summation depends on tag transition probabilities and tag emission probabilities, as well as token-level ``weights" over the tagset. These weights can be seen as defining a soft gold labeling for each token, corresponding to confidence in each label.

For clarity, define the soft gold labeling over each token $x_i$ as $\G_i \in [0,1]^{L}$, where $L$ is the size of the labelset. Now, we may define $q(.)$ as:

\[ q(\y|\x^{(s)}) = \prod_i G_i^{y_i}\]

Where $G_i^{y_i}$ is understood as the weight in $\G_i$ that corresponds to the label $y_i$.

We incorporate our instance weights in this model with the following intuitions. Recall that if an instance weight $v_i=0$, this indicates low confidence in the label on token $x_i$, and therefore the labeling should not update the model at training time. Conversely, if $v_i=1$, then this label is to be trusted entirely.

If $v_i=0$, we set the soft labeling weights over $x_i$ to be uniform, which is as good as no information. Since $v_i$ is defined as confidence in the O label, the soft labeling weight for O increases proportionally to $v_i$. Any remaining probability mass is distributed evenly among the other labels.

To be precise, for tokens in $N$, we calculate values for $\G_i$ as follows:\begin{align*}
&G_i^{O} = \max(1/L, v_i) \\
&G_i^{\text{non-}O} = \frac{1-G_i^{O}}{L-1}
\end{align*}

For example, consider phase 1 of Constrained Binary Learning, in which the labelset is collapsed to two labels ($L=2$). Assuming that the O label has index 0, then if $v_i=0$, then $\G_i = [0.5, 0.5]$. If $v_i=0.6$, then $\G_i = [0.6, 0.4]$.

For tokens in $P$ (which have some entity label with high confidence), we always set $\G_i$ with 1 in the given label index, and 0 elsewhere.

We use pretrained GloVe \cite{pennington2014glove} word vectors for English, and the same pretrained vectors used in \citet{LBSKD16} for Dutch, German, and Spanish. The other languages are distributed with monolingual text \cite{StrasselTr16}, which we used to train our own skip-n-gram vectors.

\begin{table*}
  \begin{center}
 \begin{tabular}{llrrrrrrrr|r}
\toprule
Method $\backslash$ Language & Tool & \textbf{eng} & \textbf{deu} & \textbf{esp} & \textbf{ned} & \textbf{amh} & \textbf{ara} & \textbf{hin} & \textbf{som} & avg \\
\midrule
\multirow{2}{*}{Gold} & Cogcomp & 89.1 & 72.5 & 82.5 & 82.6 & 67.2 & 53.4 & 74.4 & 80.3 & 75.3 \\
& BiLSTM-CRF & 90.3 & 77.3 & 85.2 & 81.1 & 69.2 & 52.8 & 73.8 & 82.3 & 76.5 \\
\midrule
\multirow{2}{*}{Oracle Weighting} & Cogcomp & 83.7 & 65.7 & 76.2 & 76.4 & 54.3 & 42.0 & 56.3 & 68.5 & 65.4 \\
& BiLSTM-CRF & 87.8 & 70.2 & 78.5 & 70.4 & 60.4 & 43.4 & 57.6 & 73.2 & 67.7 \\
\midrule
%\multicolumn{2}{l}{Class Weighting \cite{LDLLY03}} & 69.5 & 52.4 & 61.8 & 59.8 & 35.4 & 38.7 & 39.7  & 55.5 & 51.5 \\
\multicolumn{2}{l}{Noise Adaptation \textcolor{LinkColor}{(Hedderich, 2018)}} & 61.5 & 46.1 & 57.3 & 41.5 & -- & -- & -- & -- & -- \\
\multicolumn{2}{l}{Self-training \cite{JXLDL19}} & 82.3 & 65.2 & 76.3 & 65.5 & 52.1 & 40.1 & 55.1 & 65.3 & 62.7 \\

\midrule

\multirow{2}{*}{Raw Annotations} & Cogcomp & 54.8 & 36.9 & 49.5 & 47.9 & 31.0 & 32.6 & 30.9 & 44.0 & 40.9 \\

& BiLSTM-CRF & 73.3 & 57.7 & 61.9 & 58.3 & 42.2 & 36.8 & 47.5 & 54.9 & 54.1 \\

\multirow{2}{*}{CBL-Raw} & CogComp & 74.7 & 63.0 & 68.7 & 67.0 & 45.0 & 37.8 & 50.6 & 67.9 & 59.3 \\
& BiLSTM-CRF & \textbf{84.6} & \textbf{67.9} & \textbf{79.6} & 70.0 & \textbf{52.9} & 42.1 & 55.2 & \textbf{70.4} & \textbf{65.3} \\

\midrule

\multirow{2}{*}{Combined Weighting} & Cogcomp & 75.2 & 56.6 & 70.8 & 70.8 & 46.5 & \textbf{44.1} & 57.5 & 60.2 & 60.2 \\
& BiLSTM-CRF & 73.5 & 60.3 & 64.9 & 61.9 & 48.0 & 38.0 & 49.0 & 56.6 & 56.5 \\

\multirow{2}{*}{CBL-Combined} & Cogcomp & 77.3 & 61.8 & 74.0 & \textbf{72.4} & 49.2 & 43.7 & \textbf{58.2} & 67.6 & 63.0 \\
& BiLSTM-CRF & 81.1 & 64.9 & 74.9 & 63.4 & 52.2 & 39.8 & 52.0 & 67.0 & 61.9  \\

\bottomrule
\end{tabular}
  \end{center}
  \caption{F1 scores on English, German, Spanish, Dutch, Amharic, Arabic, Hindi, and Somali. Each section shows performance of both Cogcomp (non-neural) and BiLSTM (neural) systems. \textit{Gold} is using all available gold training data to train. \textit{Oracle Weighting} uses full entity knowledge to set weights on $N$. The next section shows prior work, followed by our methods. The column to the farthest right shows the average score over all languages. Bold values are the highest per column. On average, our best results are found in the uninitialized (\textit{Raw}) CBL from BiLSTM-CRF.}
  \label{alllangsbigtable}
\end{table*}

\subsection{Baselines}
\label{sec:weightingschemes}

We compare against several baselines, including two from prior work. 

\subsubsection{Raw annotations}
The simplest baseline is to do nothing to the partially annotated data and train on it as is. 
\subsubsection{Instance Weights}

Although CBL works with no initialization (that is, all tokens with weight 1), we found that a good weighting scheme can boost performance for certain models. We design weighting schemes that give instances in $N$ weights corresponding to an estimate of the label confidence.\footnote{All elements of $P$ always have weight 1} For example, non-name tokens such as \textit{respectfully} should have weight 1, but possible names, such as \textit{Russell}, should have a low weight, or 0. We propose two weighting schemes: frequency-based and window-based.

For the frequency-based weighting scheme, we observed that names have relatively low frequency (for example, \textit{Kennebunkport}, \textit{Dushanbe}) and common words are rarely names (for example \textit{the}, \textit{and}, \textit{so}). We weigh each instance in $N$ according to its frequency.
\begin{equation}
  v^\text{freq}_i = freq(x_i)
\end{equation}

where $freq(x_i)$ is the frequency of the $i^{th}$ token in $N$ divided by the count of the most frequent token. In our experiments, we computed frequencies over $P+N$, but these could be estimated on any sufficiently large corpus. We found that the neural model performed poorly when the weights followed a Zipfian distribution (e.g. most weights very small), so for those experiments, we took the log of the token count before normalizing.

For the window-based weighting scheme, noting that names rarely appear immediately adjacent to each other in English text, we set weights for tokens within a window of size 1 of a name (identified in $P$) to be $1.0$, and for tokens farther away to be $0$.\begin{equation}
  v^\text{window}_i =
  \begin{cases}
    1.0 & \text{if $d_i \leq 1$} \\
    0.0 & \text{otherwise}
  \end{cases}
\end{equation}

where $d_i$ is the distance of the $i^{th}$ token to the nearest named entity in $P$.

Finally, we combine the two weighting schemes as: \begin{equation}
  v^{\text{combined}}_i =
  \begin{cases}
    1.0 & \text{if $d_i \leq 1$} \\
    v^\text{freq}_i & \text{otherwise}
  \end{cases}
\end{equation}

\subsubsection{Self-training with Marginal CRF}
\citet{JXLDL19} propose a model based on marginal CRF \cite{TKOMM08} (described in Section~\ref{sec:neural}). They follow a self-training framework with cross-validation, using the trained model over all but one fold to update gold labeling distributions in the final fold. This process continues until convergence. They use a partial-CRF framework similar to ours, but taking predictions at face value, without constraints. 
 
\subsubsection{Neural Network with Noise Adaptation}
Following \citet{HedderichKl18}, we used a neural network with a noise adaptation layer.\footnote{The code was kindly provided by the authors.} This extra layer attempts to correct noisy examples given a probabilistic confusion matrix of label noise. Since this method needs a small amount of labeled data, we selected 500 random tokens to be the gold training set, in addition to the partial annotations.

As with our BiLSTM experiments, we use pretrained GloVe word vectors for English, and the same pretrained vectors used in \citet{LBSKD16} for Dutch, German, and Spanish. We omit results from the remaining languages because the scores were substantially worse even than training on raw annotations. 

\subsection{Experimental Setup and Results}
\label{baselines}
We show results from our experiments in Table \ref{alllangsbigtable}. In all experiments, the training data is perturbed at 90\% precision and 50\% recall. These parameters are similar to the scores obtained by human annotators in a foreign language (see Section \ref{sec:bengali}). We evaluate each experiment with both non-neural and neural methods.

First, to get an idea of the difficulty of NER in each language, we report scores from models trained on gold data without perturbation (\textit{Gold}). Then we report results from an Oracle Weighting scheme (\textit{Oracle Weighting}) that takes partially annotated data and assigns weights with knowledge of the true labels. Specifically, mislabeled entities in set $N$ are given weight 0, and all other tokens are given weight 1.0. This scheme is free from labeling noise, but should still get lower scores than Gold because of the smaller number of entities. Since our method estimates these weights, we do not expect CBL to outperform the Oracle method. Next, we show results from all baselines. The bottom two sections are our results, first with no initialization (\textit{Raw}), and CBL over that, then with \textit{Combined Weighting}
 initialization, and CBL over that.

\subsection{Analysis}
Regardless of initialization or model, CBL improves over the baselines. Our best model, \textit{CBL-Raw BiLSTM-CRF}, improves over the \textit{Raw Annotations BiLSTM-CRF} baseline by 11.2 points F1, and the \textit{Self-training} prior work by 2.6 points F1, showing that it is an effective way to address the problem of partial annotation. Further, the best CBL version for each model is within 3 points of the corresponding \textit{Oracle} ceiling, suggesting that this weighting framework is nearly saturated.

The \textit{Combined} weighting scheme is surprisingly effective for the non-neural model, which suggests that the intuition about frequency as distinction between names and non-names holds true. It gives modest improvement in the neural model. The \textit{Self-training} method is effective, but is outperformed by our best CBL method, a difference we discuss in more detail in Section \ref{sec:difference}. The \textit{Noise Adaptation} method outperforms the \textit{Raw annotations Cogcomp} baseline in most cases, but does not reach the performance of the \textit{Self-training} method, despite using some fully labeled data. 

It is instructive to compare the neural and non-neural versions of each setup. The neural method is better overall, but is less able to learn from the knowledge-based initialization weights. In the non-neural method, the difference between \textit{Raw} and \textit{Combined} is nearly 20 points, but the difference in the neural model is less than 3 points. \textit{Combined} versions of the non-neural method outperform the neural method on 3 languages: Dutch, Arabic, and Hindi.  Further, in the neural method, \textit{CBL-Raw} is always worse than \textit{CBL-Combined}. This may be due to the way that weights are used in each model. In the non-neural model, a low enough weight completely cancels the token, whereas in the neural model it is still used in training. Since the neural model performs well in the \textit{Oracle} setting, we know that it can learn from hard weights, but it may have trouble with the subtle differences encoded in frequencies. We leave it to future work to discover improved ways of incorporating instance weights in a BiLSTM-CRF.

In seeking to understand the details of the other results, we need to consider the precision/recall tradeoff. First, all scores in the \textit{Gold} row had higher precision than recall. Then, training on raw partially annotated data biases a classifier strongly towards predicting few entities. All results from the \textit{Raw annotations} row have precision more than double the recall (e.g. Dutch Precision, Recall, F1 were: 91.5, 32.4, 47.9). In this context, the problem this paper explores is how to improve the recall of these datasets without harming the precision.

\subsection{Difference from Prior Work}
\label{sec:difference}
While our method has several superficial similarities with prior work, most notably \citet{JXLDL19}, there are some crucial differences. 

Our methods are similar in that they both use a model trained at each step to assign a soft gold-labeling to each token. Each algorithm iteratively trains models using weights from the previous steps. 

One difference is that \citet{JXLDL19} use cross-validation to train, while we follow \citet{ChangRaRo07} and retrain with the entire training set at each round. 

However, the main difference has to do with the focus of each algorithm. Recall the discussion in Section \ref{sec:binary} regarding the two possible approaches of \textbf{1}) find the false negatives and label them correctly, and \textbf{2}) find the false negatives and remove them. Conceptually, the former was the approach taken by \citet{JXLDL19}, the latter was our approach. Another way to look at this is as focusing on predicting correct tag labels (\cite{JXLDL19}) or focus on predicting O tags with high confidence (ours).

Even though they use soft labeling (which they show to be consistently better than hard labeling), it is possible that the predicted tag distribution is incorrect. Our approach allows us to avoid much of the inevitable noise that comes from labelling with a weak model.

\input{varying_the_entity_ratio.tex}

\section{Bengali Case Study}
\label{sec:bengali}

So far our experiments have shown effectiveness on artificially perturbed labels, but one might argue that these systematic perturbations don't accurately simulate real-world noise. In this section, we show how our methods work in a real-world scenario, using Bengali data partially labeled by non-speakers.

\subsection{Non-speaker Annotations}

In order to compare with prior work, we used the train/test split from \newcite{ZPWVJKM16}. We removed all gold labels from the train split, romanized it\footnote{This step is vitally important. We used \url{www.isi.edu/~ulf/uroman.html}} \cite{Hermjakob2018OutoftheboxUR}, and presented it to two non-Bengali speaking annotators using the TALEN interface \cite{MayhewRo18}. The instructions were to move quickly and annotate names only when there is high confidence (e.g. when you can also identify the English version of the name). They spent about 5 total hours annotating, without using Google Translate. This sort of non-speaker annotation is possible because the text contains many `easy' entities -- foreign names -- which are noticeably distinct from native Bengali words. For example, consider the following:

% Original Bengali:
% এবিসি'র গিলিয়ান ফিণ্ডলে আজ প্যালেস্তাইন অধীনস্থ গাজা থেকে আজ রাতে এখবর জানিয়েছেন ।

\begin{itemize}
\item \textbf{Romanized Bengali}: ebisi'ra giliyyaana phinnddale aaja pyaalestaaina adhiinastha gaajaa theke aaja raate ekhabara jaaniyyechhena .
\item \textbf{Translation}\footnote{From \url{translate.google.com}}: ABC's Gillian Fondley has reported today from Gaza under Palestine today.
\end{itemize}

\begin{table}[t!]
  \begin{center}
    \begin{tabular}{ll}
    \toprule
    Num tokens & 49K \\
    Num sentences & 2435 \\
    Num name tokens & 2326 \\
    Entity ratio & 4.66\% \\
    Num unique name tokens & 664 \\
    Annotator 1 Prec/Rec/F1 & 84/34/48 \\
    Annotator 2 Prec/Rec/F1 & 79/28/42 \\
    Combined Prec/Rec/F1 & 83/32/47 \\
    \bottomrule
    \end{tabular}
  \end{center}
  \caption{Bengali Data Statistics. The P/R/F1 scores are computed for the non-speaker annotator with respect to the gold training data.}
  \label{bengalidata}
\end{table}

The entities are Gillian Findlay, ABC, Palestine, and Gaza. While a fast-moving annotator may not catch most of these, `pyaalestaaina' could be considered an `easy' entity, because of its visual and aural similarity to `Palestine.' A clever annotator may also infer that if Palestine is mentioned, then Gaza may be present.

Annotators are moving fast and being intentionally non-thorough, so the recall will be low. Since they do not speak Bengali, there are likely to be some mistakes, so the precision may drop slightly also. This is exactly the noisy partial annotation scenario addressed in this paper. The statistics of this data can be seen in Table \ref{bengalidata}, including annotation scores computed with respect to the gold training data for each annotator, as well as the combined score.

We show results in Table \ref{bengaliresults}, using the BiLSTM-CRF model. We compare against other low-resource approaches published on this dataset, including two based on Wikipedia \cite{TsaiMaRo16,PZMNKJ17}, another based on lexicon translation from a high-resource language \cite{MayhewTsRo17}. These prior methods operate under somewhat different paradigms than this work, but have the same goal: maximizing performance in the absence of gold training data.

\begin{table}[t!]
  \begin{center}
    \begin{tabular}{lrrr}
    \toprule
    & \multicolumn{3}{c}{Test} \\
    \cmidrule{2-4}
    Scheme & P & R & F1 \\
    \midrule
    \cite{ZPWVJKM16} & - & - & 34.8 \\
    \cite{TsaiMaRo16} & - & - & 43.3 \\
    \cite{PZMNKJ17} & - & - & 44.0 \\
    \cite{MayhewTsRo17} & - & -& 46.2 \\
    \midrule
    \multicolumn{4}{c}{\textsc{BiLSTM-CRF}} \\
    Train on Gold & 71.6 & 70.2 & 70.9 \\
    Raw annotations & \textbf{73.0} & 23.8 & 35.9 \\
    Combined Weighting & 65.9 & 34.2 & 45.0 \\
    CBL-Raw & 57.8 & \textbf{47.3} & \textbf{52.0} \\
    CBL-Combined & 58.3 & 44.2 & 50.2 \\
    \bottomrule
    \end{tabular}
  \end{center}
  \caption{Bengali manual annotation results. Our methods improve on state of the art scores by over 5 points F1 given a relatively small amount of noisy and incomplete annotations from non-speakers.}
  \label{bengaliresults}
\end{table}

\textit{Raw annotations} is defined as before, and gives similar high-precision low-recall results. The \textit{Combined Weighting} scheme improves over Raw annotations by 10 points, achieving a score comparable to the prior state of the art. Beyond that, \textit{CBL-Raw} outperforms the prior best by nearly 6 points F1, although \textit{CBL-Combined} again underwhelms.  

To the best of our knowledge, this is the first result showing a method for non-speaker annotations to produce high-quality NER scores. The simplicity of this method and the small time investment for these results gives us confidence that this method can be effective for many low-resource languages.

\section{Conclusions}
We explore an understudied data scenario, and introduce a new constrained iterative algorithm to solve it. This algorithm performs well in experimental trials in several languages, on both artificially perturbed data, and in a truly low-resource situation.

\section{Acknowledgements}
This work was supported by Contracts HR0011-15-C-0113 and HR0011-18-2-0052 with the US Defense Advanced Research Projects Agency (DARPA). Approved for Public Release, Distribution Unlimited. The views expressed are those of the authors and do not reflect the official policy or position of the Department of Defense or the U.S. Government.

\vspace{0.9cm}

\bibliography{sample,ccg}
\bibliographystyle{acl_natbib}
\end{document}

%% file: varying_the_entity_ratio.tex
\begin{table}
  \begin{center}
    \begin{tabular}{lrrr}
    \toprule
    & \multicolumn{3}{c}{Avg F1}\\
    \cmidrule{2-4}
	Method $\backslash$ $b$ & 10\% & 15\% & Gold \\
    \midrule
Oracle Weighting & 65.8 & \textbf{65.9} & 65.4 \\
Raw annotations & 40.9 & 40.9 & 40.9 \\
Combined Weighting & 59.9 & \textbf{60.2} & \textbf{60.2} \\
CBL-Combined & 62.4 & 62.3 & \textbf{63.0} \\
    \bottomrule
    \end{tabular}
  \end{center}
  \caption{Experimenting with different entity ratios. Scores reported are average F1 across all languages. \textit{Gold} $b$ value refers to using the gold annotated data to calculate the optimal entity ratio. This table shows that exact knowledge of the entity ratio is not required for CBL to succeed.}
\label{tab:btest}
\end{table}

\subsection{Varying the Entity Ratio}
\label{sec:bratio}
Recall that the entity ratio is used for balancing and for the stopping criteria in CBL. In all our experiments so far, we have used the gold entity ratio for each language, as shown in Table \ref{tab:datastats}. However, exact knowledge of entity ratio is unlikely in the absence of gold data. Thus, we experimented with selecting a default $b$ value, and using it across all languages, with the Cogcomp model. We chose values of $10\%$ and $15\%$, and report F1 averaged across all languages in Table \ref{tab:btest}.

While the gold $b$ value is the best for \textit{CBL-Combined}, the  flat 15\% ratio is best for all other methods, showing that exact knowledge of the entity ratio is not necessary.

%This is most likely related to the discussion of precision and recall above. The 15\% scheme is more likely to bias models towards higher recall, therefore better balanced predictions, and higher F1.